\documentclass[notitlepage,12pt]{jedm}
\usepackage{accessibility}
\usepackage[table]{xcolor}
\usepackage{graphicx}
\usepackage{amsmath}
\usepackage{url}
\usepackage{hyperref}
\hypersetup{
  colorlinks   = true, 
  urlcolor     = blue, 
  linkcolor    = blue, 
  citecolor   = blue 
}

\begin{document}

\title{Automated Evaluation of Classroom\\Instructional Support with LLMs and BoWs:\\Connecting Global Predictions to Specific Feedback}
\date{} 

\author{{\large Jacob Whitehill}\\Worcester Polytechnic Institute\\Worcester, Massachusetts, USA\\jrwhitehill@wpi.edu \and {\large Jennifer LoCasale-Crouch}\\Virginia Commonwealth University\\Richmond, Virginia, USA\\locasalecrj@vcu.edu}

\maketitle

\begin{abstract}
With the aim to provide teachers with more specific, frequent, and actionable feedback about their teaching,
we explore how Large Language Models (LLMs) can be used to estimate ``Instructional Support'' domain scores of the CLassroom
Assessment Scoring System (CLASS), a widely used observation protocol.
We design a machine learning architecture that uses either zero-shot prompting of Meta's Llama2, and/or a classic
Bag of Words (BoW) model, to classify individual utterances of teachers' speech
(transcribed automatically using OpenAI's Whisper) for the presence 
of Instructional Support. Then, these utterance-level judgments are aggregated over a 15-min
observation session to estimate a global CLASS score. 
Experiments on two CLASS-coded datasets of toddler and pre-kindergarten classrooms indicate that
(1) automatic CLASS Instructional Support estimation accuracy  using the proposed method (Pearson $R$ up to $0.48$) approaches
human inter-rater reliability (up to $R=0.55$); (2) LLMs generally yield slightly greater accuracy than BoW for this task, though the best
models often combined features extracted from both LLM and BoW; and (3) for classifying individual utterances, there is still
room for improvement of automated methods compared to human-level judgments. Finally, (4)
we illustrate how the model's outputs can be visualized at the utterance level to provide teachers with explainable feedback on which utterances were most positively or negatively correlated with specific CLASS dimensions.
\\

{\parindent0pt
\emph{Keywords}: classroom observation, teacher feedback, machine learning,
natural language processing, large language models
}
\end{abstract}

\section{Introduction}
A perennial challenge in modern schools is to give teachers frequent, specific, and helpful feedback about their teaching. 
Such feedback is essential for teachers' professional development \cite{lesiak2021turning} and accountability, and it also supports
educational research to assess the outcomes of various interventions.
However, in most schools, teachers typically receive feedback just a few times a year from a school principal or senior
colleague. The guidance they do receive
is often given at a high level (e.g., a short summary of an entire class period),
with only few specific references to classroom interactions that could have been handled better and little actionable
feedback for how to improve. It is also subjective and can vary in quality depending on the fatigue and
other factors of the observer.

In an effort to standardize teacher evaluation and to enable more useful feedback to teachers, educational researchers
have developed various classroom observation protocols over the years, including the Mathematical Quality
Instruction \cite{hill2008mathematical},
the Protocol for Language Arts Teaching Observations \cite{grossman2014test},
and the CLassroom Assessment Scoring System (CLASS; \cite{pianta2008classroom}).
Such protocols are usually scored \emph{globally}, i.e., just a few numbers are assigned to an entire segment (typically around 15 minutes) of classroom teaching.
Researchers from learning science, team science, and education have also developed frameworks (e.g., ``accountable talk'' \cite{o2015scaling})
for promoting effective classroom discourse as well as other teaching strategies \cite{orlich2010teaching}  to promote students' deep and critical thinking.
While classroom observation and discourse analysis frameworks
are invaluable for guiding human observers in how to
observe and rate classroom interactions and instruction, they often require significant training,
are expensive to implement, and can suffer from low inter-rater reliability \cite{ho2013reliability}. Moreover,
due to the sheer volume of classroom inter-personal interaction and speech that transpire during a typical classroom session,
it is practically impossible for human observers to capture every nuance.

For the purpose of facilitating teachers' professional growth in general, and
for improving equity \cite{lesiak2021turning} in teachers' professional development opportunities in particular, it is important to develop
new ways of giving teachers more frequent, detailed, accurate, and actionable feedback.
The last few years have seen tremendous growth in 
the accuracy of automatic speech recognition systems \cite{radford2023robust}, as well as in the capability of natural language 
processing (NLP) models
to represent the meaning of words \cite{pennington2014glove} and sentences \cite{reimers2019sentence}
and to predict and generate the most likely token sequences
following a given input (GPT, Llama2 \cite{touvron2023llama}, and many more). Researchers in education, NLP,
speech recognition, learning analytics, and other fields are thus exploring how
these new computational tools can benefit students -- e.g., by generating helpful hints and explanations of math content
\cite{pardos2023learning} -- and benefit teachers -- e.g., with specific feedback about their classroom discourse \cite{wang2023chatgpt}.

In this paper, we explore how large language models (LLMs) can analyze classroom speech and automatically
predict the human-annotated scores from a validated observation protocol (specifically, the CLASS).
These predictions could be offered to teachers as either fully automated feedback, or 
within a human-in-the-loop framework in which teachers receive suggestions and can judge for themselves whether
they agree or disagree, thus providing the machine with labels to improve its predictions.
In order to understand the relative merits for automatic classroom analysis of LLMs compared to simpler NLP methods,
we compare LLMs to a classic Bag of Words (BoW) model.
The long-term goal of our work is to explore how artificial intelligence can
provide teachers with more specific, frequent, and accurate feedback about their teaching
in an unobtrusive and privacy-preserving way.

\emph{Contributions}: (1) We devise a zero-shot prompting approach to using an LLM to estimate
a \emph{globally} scored classroom observation measure (CLASS), but still providing feedback
at a temporally \emph{local} (and possibly more actionable) level. Our machine learning approach requires only weak supervision
at the global level (15-minute classroom session), not dense supervision at the utterance level.
(2) We compare both LLM, BoW, and combined LLM-BoW methods to estimate CLASS ``Instructional Support'' domain scores,
and we show in a stratified cross-validation analysis that the estimated scores correlate
similarily with scores from human experts compared to inter-rater reliability.
(3) We illustrate how temporally specific feedback from either approach (LLM or BoW) can be both explained and visualized intuitively 
for the teacher to point out the most important utterances in the transcript.

\section{Related Work}
\subsection{CLassroom Assessment Scoring System (CLASS)}
\label{sec:class}
Noting that classrooms are a highly influential context in which children learn, much effort has been invested to understand and systematically improve classroom instruction. In terms of measurement, most recent research has focused on classroom-level measures of the quality of teacher-child interactions because both theory and empirical evidence suggest that frequent, responsive, and stimulating interactions between learners and caregivers within close and supportive relationships serve as the foundation for early development \cite{mashburn2008measures,hamre2014teachers}.
The CLASS is one of the most widely used classroom observation protocols to address this and thus is the reference point for this study.
CLASS scores have been shown to be positively associated with students' vocabulary \cite{hamre2014evidence},
phonological awareness \cite{wasik2011improving}, early reading skills \cite{burchinal2010threshold}, and more.
We provide specific details about CLASS below.

The CLASS Manual defines how the protocol is structured and how it is scored:
With the CLASS, a classroom observation session is scored in terms of multiple {\em dimensions}. Each dimension reflects a particular aspect of classroom interactions and has associated {\em behavior indicators} to ``look for'' in considering the quality of that dimension in the classroom interactions. Trained CLASS observers (more details about this are given below) score each 15-min observation session by watching either a live classroom or a videorecording post-hoc, taking notes what they observe about key interactions and events, and then assigning a holistic score that estimates the total evidence that each CLASS dimension is present during the session. Each CLASS dimension is scored on a scale of 1-7.

Next, the CLASS dimensions are partitioned into a set of {\em domains}, and each domain score is obtained by
aggregating (either averaging or summing) the constituent dimension scores.
Domain scores reflect both conceptual and correlational relationships between dimensions established in prior research.
How the CLASS domains are defined depends on the students' age group (see below), but the domains that are most commonly used 
across the different versions of the protocol
are Emotional Support, Classroom Organization, and Instructional Support. The Emotional Support domain reflects the dimensions capturing the ways in which educators establish and promote a positive climate. The Classroom Organization domain includes dimensions capturing interactions related to classroom routines and procedures that guide behavior, time and attention. Lastly, the Instructional Support domain reflects the dimensions assessing the ways in which teachers effectively promote cognitive and language development.

While the approach of conceptualizing and scoring CLASS is similar across ages,
there are different
versions of the CLASS for different age groups to reflect the unique behaviors and interactions in those settings. In this study, we used 
the CLASS-Toddler (CLASS-T) and CLASS-Pre-Kindergarten (CLASS-PreK)
as they align with the age range of children in the study samples. Further,
we focus specifically on the
``Instructional Support'' domain of each tool (in CLASS-T this is actually called ``Engaged Support for Learning''),
as it is the domain most consistently linked to improved student language and cognitive outcomes and
also the area in which educators typically show the lowest skills \cite{perlman2016systematic}.
Because it is largely rooted in the classroom discourse,
it likely lends itself particularly well to automatic analysis via NLP. The Instructional Support domain comprises three dimensions, 
briefly defined here:
%
\begin{enumerate}
\item ``Language Modeling'', which describes how teachers intentionally encourage, respond to, and expand on children's language. Behavior indicators that would reflect this dimension include teachers engaging in frequent conversation, asking open-ended questions, repeating and extending language, providing self- and parallel talk, and advanced language. Contingent responding, questions that require more than a one-word response, and elaborations on language are specific observed behaviors that would reflect high quality language modeling.
\item ``Quality of Feedback'', which reflects when a teacher responds to what a child says or does in a way that pushes the child to keep thinking or trying. Indicators that would reflect this dimension include teachers' scaffolding students, engaging in feedback loops, prompting thought processes, providing information, and encouragement \& affirmations. High quality examples include observed back and forth exchanges, asking students to explain their thinking, and expanding or clarifying their comments.
\item Either ``Facilitation of Learning and Development'' (CLASS-T), or ``Concept Development'' 
(CLASS-PreK), both of which focus on teachers' involvement with children to guide and build their thinking skills. Indicators that would reflect this dimension include teachers prompting analysis and reasoning, encouraging creation and integration of ideas, and making connections of the content to the real world. High quality observed examples of this dimension include asking ``how'' and ``why'' questions, engaging in brainstorming, and relating material to students' lives.
\end{enumerate}
Certification in the CLASS requires undergoing a multi-day training seminar,
practice annotating a set of pre-scored videos, and then a certification test.
It is up to the observer to watch a classroom session carefully and apply their knowledge of these indicators, as supported by the detailed instructions from the CLASS Manual, to each session judiciously. Although CLASS has been used in a range of education studies and found useful in understanding teacher behaviors related to student growth, 
manual CLASS coding involves a high time cost that makes it infeasible to code large numbers of videos at a fine-grained temporal resolution.
New AI-based scoring mechanisms may offer ways of giving teachers more detailed feedback more frequently.

\subsection{AI for Automated Classroom Analysis}
As machine perception has become more accurate over the past decade, researchers in educational data-mining, learning analytics, 
speech processing, and other fields have explored how it can be used to characterize classroom interactions and classroom discourse,
and how such learning analytics can be presented to teachers as useful feedback. Here we summarize the most relevant work.

\citeN{kelly2018automatically}  developed an approach to identifying ``authentic'' teacher questions,
i.e., questions without a predefined answer that stimulate student discussion. Their approach
uses automatic speech recognition to create a classroom transcript, followed by
regression trees to analyze a set of 244 syntactic and semantic features, including part-of-speech tags,
dependency structures, and discourse relations \cite{olney2017assessing}. Instead of predicting authenticity of individual questions,
they estimate the proportion of authentic questions of classroom sessions.  
The accuracy of their system (Pearson correlation of predictions to human-labeled scores),
when evaluated on datasets spanning nearly 150 classrooms, was estimated to be over $0.6$.

\citeN{dai2023search} explored multimodal approaches to detecting ``negative moments'' in the classroom, 
defined as 10-sec segments of a classroom session in which the teacher's emotion or speech was indicative of
the CLASS ``Negative Climate'' dimension (which is part of the Emotional Support domain previously described).
In particular, they assessed the accuracy of various classifiers of facial expression,
auditory emotion, and text sentiment to find negative moments in a dataset of about 1000 middle and high school classrooms.
The most promising approach to finding negative moments,
according to their study, was simple keyphrase spotting --
if an utterance contained a phrase such as 
``excuse me'', ``why are you'', ``don't talk'', etc., then it was flagged as a potentially negative moment. In their
semi-automated approach, this automated keyphrase spotting was followed by manual annotation.

\citeN{zylich2020noise} trained a custom speech recognizer to recognize a small set of
keyphrases (``thank you'', ``good job'', etc.) associated with ``supportive speech'' in toddler
classrooms. They found that the counts of how often such keywords were detected
in toddler videos (the same UVA Toddler dataset that we use in our paper) correlated, albeit to
only modest degree (the largest Pearson correlation magnitude was $0.237$), with dimensions of the human-coded CLASS scores.

\citeN{demszky2021measuring} developed an automated tool to  identify conversational
``uptake'' in school classrooms, i.e., moments when the teacher revoices, elaborates on, or asks a follow-up question
to a student contribution. In particular, they  explored different approaches
to  analyzing a pair of utterances $(S,T)$
(where $S$ is a student  utterance and $T$ is a teacher utterance) so as to predict whether $T$ directly followed $S$ in the
conversation, or whether it was chosen randomly. If the model predicts that $T$ follows $S$ directly,
then it is likely that $T$ is an example of ``uptake'' of $S$.
For classification, the authors compared (a) distance-based measures
of semantic similarity of the two utterances using GloVe \cite{pennington2014glove} or Sentence-BERT \cite{reimers2019sentence};
(b) simple proportions of the number of tokens appearing in both utterances; and (c)
a fine-tuned BERT-based \cite{devlin2018bert} classifier that takes $S$ and $T$ as input. The fine-tuned BERT worked the best.
Since the ground-truth label of whether $T$ follows $S$ can be inferred automatically from the transcript (self-supervision),
the  methodology is highly scalable. In a comparison, they found that a fully supervised approach (which requires
laborious and dense utterance-wise labeling of ``uptake'') performed only slightly better.
In later work \cite{demszky2023can}, their research team deployed the BERT-based tool in an online course on computer programming
and found that teachers, after receiving visualized feedback on their conversational uptake, increased the
degree to which they took up students' contributions.

Suresh et al.~\cite{suresh2021using} compared both LSTM and fine-tuned BERT-based models for
``TalkMove'' classification of pairs of utterances (student, teacher). TalkMoves are a framework
for analyzing classroom discourse during math instruction. Specific TalkMoves include ``keeping
everyone together'', ``restating'', ``press for reasoning'', and more.

Most similar to our work, \citeN{wang2023chatgpt} used GPT-3.5 in a zero-shot prompting framework to estimate 
CLASS scores, identify specific utterances associated with high or low performance within each CLASS dimension,
and provide suggestions for how the teachers could elicit student reasoning. 
In their task formulation, the entire transcript of a classroom session, along with a prompt about the CLASS dimension,
was fed to the language model, and the model then provided an estimate of the score. On a dataset of
elementary school math classrooms, they found that the 
Spearman correlation between the estimated and expert-labeled scores was low, i.e., between $-0.05$ and $0.07$
for the ``Instructional Dialogue'' dimension (one of the Instructional Support dimensions
in the CLASS Upper-Elementary protocol). Methodologically, our work differs from theirs in several ways:
(1) our method analyzes the classroom transcript at the utterance level instead of globally;
(2) we ask the LLM to score each utterance in terms of the CLASS-defined
indicators of each dimension, rather than directly at the dimension level;
(3) we connect global dimension scores to individual utterances via linear regression.

\emph{Outlook}: As speech recognition and other machine perception technologies
become more accurate, it is useful to evaluate whether the accuracy
of downstream classroom analysis tools also increases. Moreover, as fundamentally new technologies such as LLMs
emerge, it is exciting to explore how they can be used to enable better classroom feedback. 
Some of the challenges when applying LLMs to classroom discourse are that (1) classroom transcripts are long,
spanning thousands of tokens, which can be challenging for LLMs; (2) it is unclear how to define the task (zero-shot? few-shot?),
and to design effective prompts,
so that an LLM can make a semantically very high-level judgment about classroom discourse quality, but still connect
the \emph{globally} scored measure such as the CLASS, or the Mathematical Quality of Instruction (MQI), to \emph{specific} utterances; (3) the first work
\cite{wang2023chatgpt} on LLMs for CLASS and MQI score estimation showed mostly low-to-modest correlations with human scores.
On the other hand, recent studies by \cite{demszky2021measuring} and \cite{dai2023search} have found that
even very simple techniques such as keyword spotting can be surprisingly competitive in terms of prediction accuracy.
It is thus important to compare new generative models to classical methods to understand their relative merits.

\section{Architecture}
\begin{figure}
\begin{center}
\includegraphics[width=\textwidth]{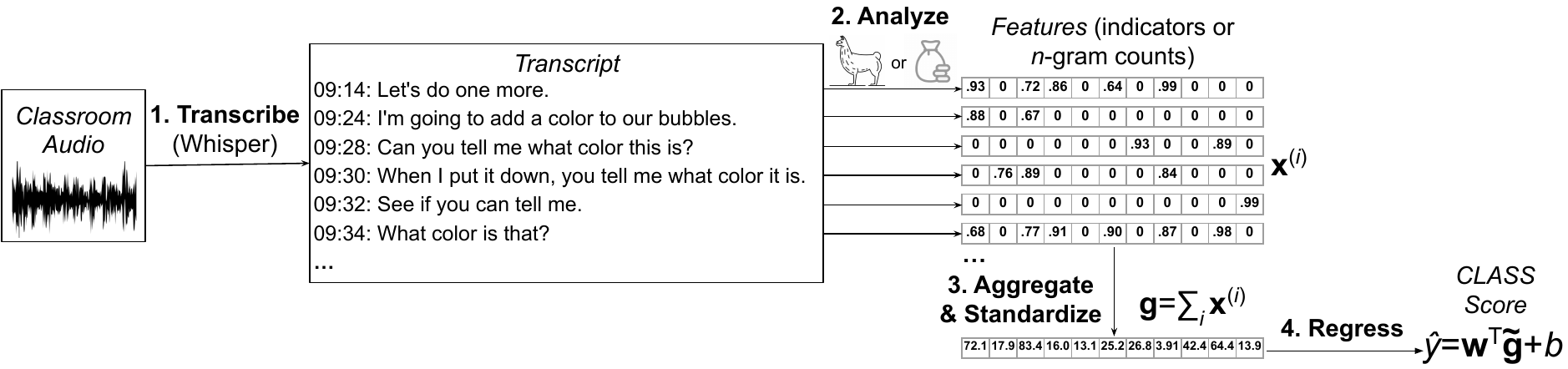}
\end{center}
\caption{Our high-level approach for CLASS score estimation: After transcribing
each classroom audio into text, each utterance is analyzed
using either Llama2 or a BoW model to produce a 
feature vector.
The feature vectors are then aggregated (summed and z-scored) across utterances, and finally regressed into a CLASS score estimate.
}
\label{fig:overview}
\end{figure}

\label{sec:approach}
We seek to develop an automatic system (see Figure \ref{fig:overview}) for analyzing a classroom transcript, which 
was produced automatically by an automatic speech recognition system, and estimating
the CLASS ``Instructional Support'' domain scores. Such a system could be used by teachers at their
own discretion, and conceivably on their own school's computer rather than a cloud-computing platform,
to obtain objective feedback about the quality of their classroom discourse.
To ensure simple explainability
and also to provide more actionable feedback, we require that the system's \emph{global} (i.e.,
over a 15-min session) estimates be tied to \emph{specific utterances} in the transcript. Moreover,
to reduce the data annotation effort, we will use only weak supervision and require just
global labels (CLASS scores), not utterance-level labels.
To aggregate the machine's judgments from individual utterances
to the entire classroom observation session, we use a simple linear model with $L_1$ regularization.

\emph{Remarks}: (1) An alternative strategy would be first to summarize the transcript and then to compute features (rather than
the reverse order). Due both to the fact that CLASS is coded based on specific events and interactions rather than at
a gestalt level, and to our goal of providing specific and actionable feedback, we chose to analyze individual utterances first
and then to aggregate afterwards (see Section \ref{sec:experiments} for a comparison with analyzing sequences of 3 utterances at a time).
(2) 
When the regression coefficient estimates themselves are of primary interest (e.g., to
compute confidence intervals around the estimates of the individual feature coefficients), then it is
standard practice to verify that the usual assumptions (homoscedasticity, linearity, etc.) of linear regression hold true,
and in particular to verify the linearity of the target variable in each of the
different covariates/features.
In our case, however, the main focus is on predictive accuracy of the model's outputs rather than
the coefficients themselves, and hence we omit this step (see 
Section \ref{sec:limitations} on Limitations). (3)  
Compared to the more commonly used ridge regression (i.e., with $L_2$ regularization), $L_1$ yields
regression coefficients that are sparse (i.e., for most features they are 0), which affords greater model
interpretability.
(4) We also considered \emph{non-negative} $L_1$-regularized
linear regression to enforce non-negativity of all feature coefficients (see Appendix).


Given this high-level design, the key question is how to make reliable judgments of
individual utterances. We compare two methods:  (1) Large Language Models (LLMs), which have
demonstrated impressive ability to make accurate semantic inferences on a wide variety of tasks and may also hold significant promise for
classroom feedback; and (2) classic Bag of Words (BoW) models, which
have the advantage of determinism/reproducibilty, efficiency/speed, and simplicity of implementation. Moreover,
findings by \cite{dai2023search} and \cite{demszky2021measuring} suggest that such simple methods can be surprisingly effective.

\subsection{Datasets}
\label{sec:datasets}
Before describing the two methods in detail, we first present the two datasets we used for training and evaluation in our experiments.
\begin{enumerate}



\item \emph{UVA Toddler}:
The University of Virginia (UVA) Toddler dataset \cite{locasale2023results,locasale2016unpacking} 
consists of 172 classroom observation sessions (each 15 minutes long) from 
classrooms serving children 18-36 months (with 47 teachers in total) in a south-eastern state of the United States. Teacher demographics:
50\% Black/African-American, 38\% White/Caucasian, 4\% Asian, 2\% Latino/Hispanic/Spanish, 2\% Multiracial, and 4\% Other. All teachers were female.
No child demographic information is available in the dataset description by \cite{locasale2023results}.
The sessions were scored
for the CLASS-Toddler protocol by a team of 9 annotators, and on average each 15-min CLASS session was scored by
1.30 different annotators. The average transcript length (obtained automatically from Whisper)
is 1204.34 words per session.
\item \emph{NCRECE PreK}:
The National Center for Research on Early Childhood Education Pre-Kindergarten (NCRECE PreK) dataset \cite{ncrece,pianta2017early}
consists of several thousand videos from
pre-kindergarten classrooms across nine socio-demographically diverse U.S. cities in the United States.
In our study, we utilized a random subset of NCRECE PreK sessions consisting
of 561 classroom sessions (each 15 minutes long) from 41 unique teachers.
In particular, the subset was chosen by selecting an arbitrary video from the entire database, and then adding
all CLASS-coded sessions from that video to the subset; see Appendix for more details.
Teacher demographics of the sample:
55\% Black/African-American, 31\% White/Caucasian, 12\% Multiracial, and 2\% Other. 93\% of the teachers were female.
Child demographics of the sample:  50\% Black/African-American, 29\% Hispanic, 13\% White/Caucasian, 4\% Asian, 3\% Multiracial, and 1\% Other.
52\% of the children were female.
These sessions were scored by a team of 43 annotators, and on average each 15-min CLASS session was scored by
1.67 different annotators. The average transcript length
is 1585.51 words per session.



\end{enumerate}
The human inter-rater reliabilities of the CLASS labels for these datasets are given in Table \ref{tbl:results}.

\subsubsection{Preprocessing}
Each video was transcribed automatically using OpenAI's Whisper \begin{tt}large-v2\end{tt} automatic speech recognition
model \cite{radford2023robust}, which is state-of-the-art. Whisper estimates the start and end times of each detected utterance.
Given the student age groups (toddler and pre-kindergarten) in our datasets, as well as the fact that the children's
speech tended to be quieter and less clear than that of the adults, most of the detected speech came from the teachers
rather than the students.\footnote{
While it would be conceivable to try to detect segments of children's speech and then somehow filter it
to boost the signal-to-noise ratio, we followed standard practice and applied Whisper to the raw speech signal.
} Nonetheless, the teachers' speech still captures rich teacher-student interactions, e.g.,
by prodding students to think more deeply (``Why do you think she ran away?''),
repeating or rephrasing what a student said (``You say you drew a circle?''), or narrating a teacher's reaction to something the student did (e.g., ``I like how you are coloring, Jennifer.'').
We removed those sessions that contained no detected speech.
Note that Whisper occasionally splits a single sentence (``I wish you would stop speaking so loudly.'')
into multiple sentence fragments (``I wish you would'' and ``stop speaking so loudly.''). We decided not to try to merge
such speech and left them as-is. Also, it is possible that Whisper's accuracy 
differs across different demographic groups in our dataset; since we do not have ground-truth transcripts, we cannot
assess this possible differential accuracy. 

\subsubsection{Inter-Rater Reliability}
\label{sec:irr}
To ensure the greatest possible comparability with the automated predictors,
we assessed human inter-rater reliability (IRR) of the CLASS scores in a leave-one-labeler-out fashion. Specifically, we computed
the average (over all $k$ labelers) Pearson correlation $R$,
as well as the average root-mean-squared error (RMSE), between each labeler's labels and the mean labels assigned to the sessions
by all the \emph{other} labelers.
We also report the standard error of these averages, estimated as the standard deviation divided by $\sqrt{k}$.
In this way, we use the same metrics to quantify human IRR as we do for the accuracy of the automated methods.

\subsection{Method I: Large Language Models (LLM)}
\label{sec:indicators}

In our experiments we 
used Meta's Llama2-7b-chat model \cite{touvron2023llama} to assess each input utterance for multiple indicators pertaining to
the CLASS ``Instructional Support'' domain. With the Llama2 chat models, both a \begin{tt}system\end{tt} message
and a \begin{tt}user\end{tt}  message are input: the former can be used to provide general instructions for \emph{how} the LLM should respond,
and the latter contains the specific queries to which it should respond. Our general approach to applying Llama2 was to ask
it to analyze an individual utterance from a classroom transcript and infer whether or not the utterance exhibits
one of the behavioral indicators (see Section \ref{sec:class}) associated with 
CLASS Instructional Support. Accordingly, we gave Llama2 a 
\begin{tt}system\end{tt} message of \begin{tt}"Answer YES or NO."\end{tt} and a \begin{tt}user\end{tt} message of:
\footnotesize
\begin{verbatim}
"In the context of a preschool classroom in which a teacher is talking to
their students, does the following sentence '<indicator>' and help students
to grow cognitively?\n<input text>"
\end{verbatim}
\normalsize
Here, \begin{tt}<input text>\end{tt} was
a single detected utterance of classroom speech. For the NCRECE PreK dataset (labeled for CLASS-PreK), the \begin{tt}<indicator>\end{tt}
was one of the following:
        (1) ``promote analysis and reasoning'',
        (2) ``facilitate creativity by brainstorming and/or planning'',
        (3) ``help students to make connections'',
        (4) ``provide scaffolding'',
        (5) ``provide information'',
        (6) ``ask students to explain their reasoning'',
        (7) ``encourage and affirms'',
        (8) ``ask open-ended questions'',
        (9) ``repeat and extend students' language'',
        (10) ``perform self- and parallel talk'', and
        (11) ``use advanced language''.
Indicators (1)-(3) correspond to ``concept development'',
indicators (4)-(7) to ``quality of feedback'', and (8)-(11) to ``language modeling''.
For the UVA Toddler dataset, the prompts were changed slightly to match the indicators of the CLASS-T
dimensions; see Appendix. 
Below is an example of a chat sent to Llama2 for the input text, ``What animal roars?'':
\footnotesize
\begin{verbatim}
{"role": "system", "content": "Answer YES or NO."},
{"role": "user", "content": """In the context of a preschool classroom in
which a teacher is talking to their students, does the following sentence
'promote analysis and reasoning' and help students to grow cognitively?
"What animal roars?" """ }
\end{verbatim}
\normalsize

After feeding the prompt to Llama2, we then parsed its response to determine whether or not the first token of
the response was the ``YES'' token.
If it was, then we obtained from Llama2 the probability of this token given the prompt (by enabling the \begin{tt}logprobs\end{tt}
flag in Llama2's \begin{tt}generator.chat\_completion\end{tt} method) and used it as a feature value. If it was not, then we set the feature value to 0.\footnote{This is actually an underestimate,
since the probability of ``YES'' is positive even if it is not the most likely token, but it simplified the processing since
Llama2 only outputs log-probabilities for the most likely token.}
Our intuition was that Llama2's \emph{confidence} in ``YES'' -- not just the binary presence/absence -- could contain useful information about the
queried indicator in the input sentence.
Across the 11 different indicators, this approach formed a real-valued vector ${\bf x}^{(i)} \in [0,1]^{11}$
for each utterance $i$. We processed each classroom transcript in mini-batches, where one mini-batch consisted of 11 chats (corresponding
to the 11 indicators) for a single utterance. Note that we also tried some alternative LLM-based approaches; see Appendix.

\subsection{Method II: Bag of Words (BoW)}
We employed a Bag of Words (BoW) approach to analyze the transcript of each 15-minute classroom session. We performed
this utterance-wise to compute the number of occurrences of each of a fixed set of words within each utterance.
As a preprocessing step, we first converted all the  utterances detected by Whisper to lower-case and then removed commas, periods, and a set of
stop-words (see Appendix).
For the set of ``words'', we considered both individual words as well
as word sequences (known as ``shingling'' \cite{manning2010}):
Across all utterances over all videos within each dataset, we computed the set of unique word sequences ($n$-grams)
within each utterance, where the sequence length $n \in \{1,2,3,4\}$.
From this set of possible $n$-grams, we selected  the
$300$ most frequently elements (separately for each dataset; see Appendix) 
and then manually added two more ``words'' consisting of '?' and ' ', which
implicitly enabled the BoW model to count the number of questions and total number of words within each transcript.
Next, we extracted a $302$-dim feature vector from each detected utterance from each classroom video,
consisting of the counts (within each utterance) of each of the 300 selected $n$-grams and the two manually added '?' and ' ' words.

\subsection{Feature Aggregation and $L_1$ Regression to Estimate CLASS Scores}
\label{sec:regression}
The feature vector for each utterance consisted of either the LLM probability for each behavioral indicator
or the BoW count for each $n$-gram. 
We  then aggregated the feature vectors over all utterances in each transcript, z-scored the result, and then regressed
the CLASS score.
Let ${\bf x}^{(i)}$ be the feature vector of utterance $i$ in a classroom observation session.
We first compute a global feature vector for the session as ${\bf g} = \sum_{i=1} {\bf x}^{(i)}$, where the summation
is over all utterances in the session.
We chose to sum (rather than, say, average) the utterance-wise feature vectors since CLASS scores are intended
to represent the \emph{total} amount of evidence for each dimension (rather than, say, the proportion of the session that exhibits
the dimension). Next, we z-scored the elements of ${\bf g}$ by subtracting
the mean feature vector ${\bf m}$ and dividing by the standard deviation ${\bf s}$, where ${\bf m}$ and ${\bf s}$
are computed over the entire training set of classroom sessions. This yields
a standardized global feature vector ${\bf \tilde g} = ({\bf g} - {\bf m}) / {\bf s}$ where the 
division is computed element-wise.
Finally, we predict the CLASS scores for the session as
$\hat{y} = {\bf w}^\top {\bf \tilde g} + b$, where ${\bf w}$ are the regression weights and $b$ is the
bias term.

\section{Experiments}
\label{sec:experiments}
Using the datasets described in Section \ref{sec:datasets},  we conducted experiments to compare the accuracy 
of the LLM and BoW approaches for CLASS score prediction.
Experiments were conducted on an NVIDIA A100 GPU with 40GB of RAM. We used the official Open AI and Meta code repositories for
Whisper and Llama2, respectively.
As accuracy metrics, we used both the root mean squared error (RMSE) and Pearson's correlation coefficient $R$. We prefer Pearson
correlations (rather than Spearman rank correlation or Cohen's $\kappa$) for two reasons: (1)  We
are quantifying the accuracy of a continuous-valued estimator $\hat{y}$ with respect to another continuous-valued target (the average
CLASS score of each session from multiple human annotators). (2) It is arguably more useful to develop a predictor
whose scores $\hat{y}$ are linearly related to the target value $y$, not just monotonically related.
Note that Pearson correlation 
measures the ability of a predictor to explain the variance in CLASS scores above-and-beyond always predicting the central tendency
of the test set. $L_1$-regularized regression models from feature vectors to CLASS scores were trained
for each of the 3 separate CLASS dimensions as well as the entire CLASS domain (i.e., the sum of the three dimensions' scores).
Moreover, for the LLM approach, we tried subselecting elements of the feature vector that were related to only a specific
dimension. For instance, the ``LLM(ConDev)'' feature vector includes the LLM's judgments for only those
indicators pertaining to the Concept Development dimension (see Section \ref{sec:indicators}); this allowed us to investigate
whether the LLM can reason specifically about particular aspects of the CLASS rather than to just the Instructional Support
domain as whole. The ``LLM(All)'' feature vector consists of all 11 behavioral indicators.
To test whether the BoW and LLM approaches are complementary,
we also tried combining the feature vectors of both the LLM and BoW approaches (``LLM(All) $\|$ BoW'').
Finally, we also  implemented several baselines consisting of feature vectors with counts of the number of words and/or
the number of questions in each transcript.

\emph{Cross-validation}:
To measure generalization accuracy, we used 5-fold cross-validation, constructed so that no teacher appeared in more than one fold and
so that each fold spanned a range of small to large CLASS scores. We set the $L_1$ regularization strength to $0.1$,
and the Llama2 temperature and top-$p$ probability to $0.6$ and $0.9$, respectively, for
all models in both datasets. Each model was trained on all the videos belonging to the teachers in the training fold and
then evaluated on all the videos belonging to the teachers in the testing fold. 
We computed the average (across the 5-folds) of the Pearson correlations ($R$) and 
root mean squared errors (RMSE), along with standard error estimates (standard deviation across the folds divided by
$\sqrt 5$, as suggested by  \cite{tibshirani2014}; we acknowledge, however, that no unbiased estimate of the variance exists for cross-validation \cite{bengio2003no}). We also report inter-rater reliability of human annotators for comparison.

\emph{Model Variations}: Finally, we explored several variations to the LLM-based approaches described in Section \ref{sec:indicators}:
\begin{enumerate}
\item Binary features: instead of the real-valued probability that the first word of the 
LLM's response was ``YES'', we tried binarizing the result.
\item Larger LLM: instead of the Llama2-7b-chat model, we tried the larger Llama2-13b-chat model to 
see if larger LLMs might yield higher accuracy.
\item 3-sentence context: instead of analyzing each utterance $i$ individually for the presence of the 11 CLASS
indicators, we analyzed the concatenation of the utterances $(i-1,i,i+1)$.
For the very first (last) utterance in the transcript, we defined the preceding (succeeding) utterance to
be the empty string.
\end{enumerate}
For each variation, we trained the LLM(All) model and assessed its accuracy on the overall Instructional Support
domain for each dataset.

\subsection{Results}
\begin{table}
\caption{Prediction accuracy (using 5-fold cross-validation), as expressed in terms of Pearson correlation $R$ and RMSE,
on the UVA Toddler (${\bf top}$) and NCRECE PreK (${\bf bottom}$) datasets. Standard error estimates are given in parentheses.
$\|$ indicates the concatenation of multiple feature vectors.
Bold-face results are the best automated prediction accuracy within each column.}
\label{tbl:results}
\setlength\tabcolsep{2.0pt} 
\footnotesize
\begin{tabular}{|l||c|c|c|c|c|c||c|c|}\hline
\multicolumn{9}{|c|}{\bf UVA Toddler} \\ \hline 
             & \multicolumn{2}{c|}{\bf Fac Learn \& Dev} & \multicolumn{2}{c|}{\bf Qual Fdbk} & \multicolumn{2}{c||}{\bf Lang Modeling} & \multicolumn{2}{c|}{\bf Inst~Support}  \\ \hline 
{\bf Method} & {\bf R$\uparrow$} & {\bf RMSE$\downarrow$} & {\bf R$\uparrow$} & {\bf RMSE$\downarrow$} & {\bf R$\uparrow$} & {\bf RMSE$\downarrow$} & {\bf R$\uparrow$} & {\bf RMSE$\downarrow$} \\ \hline 
\multicolumn{9}{|c|}{\em Inter-Rater Reliability} \\ \hline 
Human labelers & 0.53 (0.10) & 1.24 (0.10) & 0.24 (0.22) & 0.80 (0.19) & 0.32 (0.21) & 1.29 (0.14) & 0.37 (0.20) & 2.40 (0.21) \\ \hline 
\multicolumn{9}{|c|}{\em Baselines} \\ \hline 
\#words & 0.27 (0.10) & 1.15 (0.08) & 0.23 (0.12) & 1.23 (0.09) & 0.21 (0.11) & 1.33 (0.03) & 0.26 (0.12) & 3.20 (0.18)\\ \hline
\#questions & {\bf 0.34 (0.07)} & 1.12 (0.08) & 0.28 (0.09) & 1.21 (0.10) & 0.28 (0.07) & 1.30 (0.04) & 0.34 (0.09) & 3.11 (0.21)\\ \hline
\#words $\|$ \#ques. & 0.30 (0.08) & 1.13 (0.08) & 0.22 (0.10) & 1.23 (0.10) & 0.26 (0.08) & 1.30 (0.04) & 0.30 (0.10) & 3.15 (0.21)\\ \hline
\multicolumn{9}{|c|}{\em Proposed Methods} \\ \hline 
BoW & 0.31 (0.11) & 1.24 (0.15) & 0.33 (0.08) & 1.23 (0.09) & 0.31 (0.04) & 1.41 (0.10) & {\bf 0.39 (0.08)} & 3.72 (0.44)\\ \hline
LLM(FacLDev) & 0.27 (0.07) & 1.14 (0.08) & 0.31 (0.10) & 1.20 (0.09) & 0.25 (0.08) & 1.31 (0.04) & 0.31 (0.08) & 3.16 (0.18)\\ \hline
LLM(QualFbk) & 0.32 (0.07) & {\bf 1.12 (0.08)} & 0.35 (0.12) & {\bf 1.18 (0.10)} & {\bf 0.33 (0.08)} & {\bf 1.27 (0.02)} & 0.38 (0.10) & {\bf 3.04 (0.19)} \\ \hline
LLM(LangMod) & 0.27 (0.07) & 1.18 (0.09) & 0.28 (0.09) & 1.26 (0.10) & 0.19 (0.08) & 1.38 (0.08) & 0.26 (0.08) & 3.43 (0.30)\\ \hline
LLM(All) & 0.25 (0.08) & 1.17 (0.09) & 0.34 (0.11) & 1.22 (0.10) & 0.29 (0.09) & 1.30 (0.05) & 0.32 (0.10) & 3.25 (0.25)\\ \hline
LLM(All) $\|$ BoW & 0.32 (0.11) & 1.18 (0.12) & {\bf 0.35 (0.09)} & 1.21 (0.09) & 0.30 (0.04) & 1.40 (0.09) & 0.39 (0.08) & 3.70 (0.41)\\ \hline
\end{tabular}

\vspace{.5cm}

\begin{tabular}{|l||c|c|c|c|c|c||c|c|}\hline
\multicolumn{9}{|c|}{\bf NCRECE PreK} \\ \hline 
             & \multicolumn{2}{c|}{\bf Con Dev} & \multicolumn{2}{c|}{\bf Qual Fdbk} & \multicolumn{2}{c||}{\bf Lang Modeling} & \multicolumn{2}{c|}{\bf Inst~Support}  \\ \hline 
{\bf Method} & {\bf R$\uparrow$} & {\bf RMSE$\downarrow$} & {\bf R$\uparrow$} & {\bf RMSE$\downarrow$} & {\bf R$\uparrow$} & {\bf RMSE$\downarrow$} & {\bf R$\uparrow$} & {\bf RMSE$\downarrow$} \\ \hline 
\multicolumn{9}{|c|}{\em Inter-Rater Reliability} \\ \hline 
Human labelers & 0.49 (0.04) & 1.25 (0.05) & 0.35 (0.07) & 1.22 (0.08) & 0.44 (0.04) & 1.25 (0.05) & 0.55 (0.04) & 2.79 (0.15) \\ \hline 
\multicolumn{9}{|c|}{\em Baselines} \\ \hline 
\#words & 0.23 (0.04) & 1.03 (0.04) & 0.16 (0.06) & 0.98 (0.03) & 0.19 (0.05) & 1.01 (0.04) & 0.22 (0.05) & 2.60 (0.09)\\ \hline
\#questions & 0.26 (0.05) & 1.02 (0.04) & 0.20 (0.06) & 0.97 (0.03) & 0.21 (0.05) & 1.01 (0.04) & 0.26 (0.06) & 2.57 (0.09)\\ \hline
\#words $\|$ \#ques. & 0.26 (0.05) & 1.02 (0.04) & 0.20 (0.06) & 0.97 (0.03) & 0.21 (0.05) & 1.01 (0.04) & 0.26 (0.06) & 2.57 (0.08)\\ \hline
\multicolumn{9}{|c|}{\em Proposed Methods} \\ \hline 
BoW & 0.36 (0.04) & 0.97 (0.04) & 0.23 (0.04) & 0.95 (0.04) & 0.41 (0.05) & 0.94 (0.03) & 0.47 (0.04) & 2.34 (0.08)\\ \hline
LLM(FacLDev) & 0.38 (0.04) & 0.98 (0.04) & 0.29 (0.06) & 0.95 (0.04) & 0.40 (0.05) & 0.96 (0.04) & 0.41 (0.05) & 2.46 (0.08)\\ \hline
LLM(QualFdbk) & 0.20 (0.05) & 1.04 (0.04) & 0.19 (0.04) & 0.97 (0.03) & 0.08 (0.03) & 1.02 (0.05) & 0.19 (0.05) & 2.63 (0.11)\\ \hline
LLM(LangMod) & 0.37 (0.04) & 0.99 (0.04) & {\bf 0.31 (0.05)} & 0.95 (0.04) & 0.40 (0.05) & 0.96 (0.03) & 0.41 (0.05) & 2.49 (0.08)\\ \hline
LLM(All) & 0.38 (0.04) & 0.98 (0.04) & 0.30 (0.05) & 0.95 (0.04) & 0.41 (0.05) & 0.96 (0.04) & 0.42 (0.05) & 2.47 (0.08)\\ \hline
LLM(All) $\|$ BoW & {\bf 0.41 (0.04)} & {\bf 0.95 (0.05)} & 0.27 (0.03) & {\bf 0.94 (0.04)} & {\bf 0.44 (0.04)} & {\bf 0.93 (0.03)} & {\bf 0.48 (0.04)} & {\bf 2.32 (0.08)}\\ \hline

\end{tabular}
\end{table}

Results (Pearson $R$ and RMSE) are shown in Table \ref{tbl:results} separately for each prediction task (the three CLASS dimensions
for each age group, along with the combined domain score). All numbers 
are displayed with two decimal digits of precision, along with standard error
estimates in parentheses and the
best method in each column (modulo the ``Human labelers'' row) rendered in bold.
Note that the RMSE for the CLASS domain is about 3 times higher than for each individual CLASS dimension; this is because the dimension
score was computed by summing the 3 constituent dimension scores.

\emph{CLASS-T}:
On the UVA Toddler dataset, the best automated approach for the CLASS Instructional Support domain prediction task achieved Pearson $R=0.39$.
The best automated predictors were comparable in accuracy to human IRR on all prediction tasks except
for the Facilitation of Learning and Development dimension, in which the human IRR was substantially better.
Note that IRR was quite low on the other dimensions (Pearson $R$ between $0.24-0.37$), suggesting the difficulty and/or subjectivity
of the annotation task, or possibly insufficient training of the annotators.  From the automated approaches,
there was no universally best feature vector, but the best feature vector for each task usually came from the LLM rather than from the BoW model.
Compared to both the BoW and LLM approaches, the baselines (number of words, questions, or both, within each transcript)
were surprisingly competitive; in fact, for the Facilitation of Learning and Development dimension, the \#questions baseline achieved the
highest Pearson correlation with human annotations.
For the baselines,
the $L_1$ regression coefficients of the number of words (\#words) and/or number of questions (\#ques)
features were always positive, indicating that, in general, more classroom speech is associated with higher Instructional Support scores.

\emph{CLASS-PreK}:
On the NCRECE PreK dataset, the best automated approach for Instructional Support domain prediction was $R=0.48$.
Over each dimensions/domain, human IRR was higher (Pearson $R$ between $0.35-0.55$) than for UVA Toddler -- possibly
because there were more annotators for NCRECE PreK than UVA Toddler, and thus the ``mean label'' to which 
each annotator's scores were compared was statistically more reliable (see Section \ref{sec:irr}).
Automated prediction accuracy was also generally better
than in UVA Toddler. The combined approach ``LLM(All) $\|$ BoW'' almost always performed best, usually
just slightly more accurate than LLM(All) or the BoW model by itself. The relative benefit of LLM, and also of BoW, relative to the baselines
(which attained Pearson $R$ of at most $0.26$ on any dimension/domain) was stronger compared to in the UVA Toddler dataset.
Similar to CLASS-T, the regression coefficients of the baseline features were always positive.
The fact that the accuracy attained by the LLM approaches on NCRECE PreK was higher than on UVA Toddler
could be because the semantic judgments required for CLASS-PreK, with more advanced language,
align better with Llama2's strengths. In toddler classrooms, there are often
more ``tasks'' to do -- such as  feeding, diapering, etc. -- and the language exchanges
between teachers and students tend to be more focused on such tasks rather than expanding language and learning.  

\emph{Dimension-Specific LLMs}:
There was no evidence that the dimension-specific LLM feature vectors
were more accurate for predicting their designated dimension
than the other LLM prompts. In fact, for the 3 different CLASS dimensions on both datasets, the dimension-specific
predictor was usually not the most accurate one. This could potentially be because the different dimensions are related to each other
(since they all belong to the Instructional Support domain), or because of statistical noise induced by the relatively low IRR.

\emph{LLM vs.~BoW}:
The LLMs usually outperformed the BoW models, but not always. BoW was especially strong
for predicting Instructional Support domain scores on both UVA Toddler ($0.39$)  and NCRECE PreK ($0.47$), and on UVA Toddler
it was the best automated method for this task.
When the LLMs did outperform BoW, the
difference was often small; this is in-line with results by \cite{demszky2021measuring}.
Moreover, there was utility in combining the feature vectors from the two approaches. 




%

\emph{Overall Accuracy}:
The correlations in our experiments were generally higher than those found by \citeN{wang2023chatgpt};
in their paper, the highest reported correlation with any CLASS dimension was $0.35$ (Spearman $R$).
While the results are not directly comparable (different metric, different dataset and age group, GPT vs.~Llama2, etc.),
we suspect the main reason for the accuracy difference
is that the ``long-form'' approach used by \cite{wang2023chatgpt} -- whereby
the entire transcript was input to the LLM -- does not provide the model with enough scaffolding.
In contrast, our approach decomposes the estimation task in two ways: instead of asking the model
to predict a CLASS score for an entire dimension, we instead asked it to judge the presence/absence
of a specific indicator within a dimension. Moreover, instead of asking Llama2 to judge an entire
transcript, we instead asked it to judge just a single utterance.

\emph{Model Variations}: Results are shown in Table \ref{tbl:variations}; for comparison,
we also included the corresponding data from Table \ref{tbl:results} for the original LLM(All) models.
The only advantage over our originally proposed
model (Section \ref{sec:indicators}) was for the 3-sentence-context and only for the UVA Toddler dataset. It is
possible that the additional context was useful due to the typically very short utterance length 
in the classrooms of the younger students. The larger LLM showed no accuracy advantage on this task,
and it is also slower and/or requires more GPU hardware. Finally, the fact that the binary features
performed worse suggests that the real-valued confidence scores provided by Llama2 for each indicator
are informative.

\begin{table}
\caption{Architectural variations of the LLM model and corresponding accuracies.}
\label{tbl:variations}
\begin{center}
\begin{tabular}{|l||c|c||c|c|}\hline
& \multicolumn{2}{c||}{\bf UVA Toddler} & \multicolumn{2}{c|}{\bf NCRECE PreK} \\ \hline
{\bf Method} & {\bf R$\uparrow$} & {\bf RMSE$\downarrow$} & {\bf R$\uparrow$} & {\bf RMSE$\downarrow$}  \\ \hline
LLM(All) (from  Table \ref{tbl:results}) & 0.32 (0.09) &  3.25 (0.25) & {\bf 0.42 (0.05)} & 2.47 (0.08)  \\\hline
Binary features & 0.30 (0.09) & 3.27 (0.25)  & 0.40 (0.05)& 2.49 (0.08) \\\hline
Llama2-13b-chat & 0.31 (0.07) & 3.11 (0.19) & 0.34 (0.07) & 2.57 (0.07) \\ \hline
3-sentence context & {\bf 0.37 (0.11)} & {\bf 3.08 (0.22)} & 0.39 (0.04) & {\bf 2.46 (0.10)} \\ \hline
\end{tabular}
\end{center}
\end{table}

\emph{Other accuracy metrics}: Prediction accuracy as well as human inter-rater reliability were generally very
similar when assessed with Spearman rank correlation rather than Pearson. For brevity, we report only a few key results on the
CLASS Instructional Support domain scores: IRR was $0.35$ on UVA Toddler and $0.55$ on NCRECE.
On UVA Toddler, the best model (according to Spearman correlations) was the LLM(FacLDev), which 
achieved a Spearman correlation of $0.36$, and on
NCRECE, it was the LLM(All) $\|$ BoW model, which achieved a Spearman correlation of $0.49$.
Finally, we also report quadratically-weighted Cohen's $\kappa$, which we obtained by linearly scaling $\hat{y}$ to $[1,7]$ (for dimension
scores) or $[1,21]$ (for domain scores) and then rounding to the nearest integer: On UVA Toddler using the LLM(FacLDev) model,
it was $0.27$, and on NCRECE using the LLM(All) $\|$ BoW model, it was $0.36$.

\section{Analysis of Llama2's Judgments of Individual Utterances}
The experiments above assessed the models' ability to estimate CLASS scores at the global level of
classroom transcripts. Here, we inspect their ability to infer whether individual utterances
displayed Instructional Support. To make this idea precise, we define the \emph{marginal score} $\Delta \hat{y}^{(i)}$
of a single utterance $i$ as how much it increases the overall CLASS score estimate $\hat{y}$ of the entire transcript.
Recall that, after training the regression model to yield parameters
${\bf w}$ and $b$ as well as standardization parameters ${\bf m}$ and ${\bf s}$,
we can estimate the CLASS score as  $\hat{y} = {\bf w}^\top {\bf \tilde g} + b =
{\bf w}^\top (\sum_i {\bf x}^{(i)} - {\bf m})/{\bf s} + b$.
When we add a feature vector ${\bf x}$ corresponding to a new utterance to this sum, we obtain $\hat{y}' = 
{\bf w}^\top \left(\sum_i {\bf x}^{(i)}  + {\bf x}- {\bf m}\right)/{\bf s} + b$.
Hence, the marginal score
\[\Delta \hat{y} \doteq \hat{y}' - \hat{y} = ({\bf w}/{\bf s})^\top {\bf x}\]
where ${\bf w}/{\bf s}$ represents the standardized weights vector.

Given this definition, we investigated how much the marginal scores $\Delta \hat{y}^{(i)}$ according to the LLM(All) model
correlated with human judgments of Instructional Support.
To this end, the two authors of this paper, both of whom are CLASS-trained,
labeled a subset of the utterances in the NCRECE Pre-K dataset for whether or not each utterance
exhibited Instructional Support. The labeling was performed independently and blindly of the other labeler's 
as well as of the LLM's judgments.
In particular, a set of 100 utterances was extracted from one of the
test folds of the cross-validation procedure described in Section \ref{sec:experiments}. Using
the trained regression weights for the LLM(All) model, the utterances were selected to span the
entire range $[\min_i(\Delta \hat{y}^{(i)}), \max_i(\Delta \hat{y}^{(i)})]$ of the model's estimates of the CLASS domain scores 
on this test fold; this was achieved by sorting the $\Delta \hat{y}$ values and then picking every $(n/100)$th utterance in sequence,
where $n$ was the total number of utterances in the test fold. Then, the Pearson correlation $R$ between 
the $\Delta \hat{y}$ scores and the average of the two human labelers' scores was computed, as well as
the IRR of the two labelers themselves. In addition, to get a sense of how well the LLM(All) model can make coarse-grained
distinctions between
utterances associated with very strong vs.~very weak estimated Instructional Support, we also computed the Pearson
correlation with human judgments on the top and bottom 10 scores according to the estimated $\Delta \hat{y}$ values.

We note that manually scoring individual utterances is
not standard practice in CLASS scoring (due in part to the high labor involved), and hence it is not known
whether such coding tends to correlate highly with CLASS scores.

\subsection{Results}
The correlation between LLM(All) and the average of the two labelers' scores
on the set of 100 utterances was $0.35$. On the subset of 20 utterances corresponding to the highest and lowest
$\hat{y}$ scores, the correlation was $0.52$.
For comparison, the IRR on all 100 utterances was $0.75$. This indicates that, for the task of assessing Instructional  Support
on individual utterances, there is still substantial room for improvement compared  to human-level accuracy.


\begin{table}
\caption{Specific utterances that the LLM(All) model believed to 
exhibit particularly high or low CLASS Instructional Support.
Ellipses ($\ldots$) are inserted where the utterance began/ended abruptly due to how Whisper segmented the transcript.
}
\label{tbl:utterances}
\footnotesize
\begin{center}
\begin{tabular}{|l||l|}
\multicolumn{2}{c}{\bf Sample Utterances with Highest/Lowest Estimated Inst.~Supp.} \\ \hline 
{\bf High Inst.~Supp.} & {\bf Low Inst.~Supp.} \\ \hline 
Malia said maybe they have to cry. So if they're hungry and & Aria, give me an animal that was in the story.\\
they want a bottle, how do you think their mommy knows? & \\ \hline
Well, let's find out. & Oh, maybe you should go look for the letter J. \\\hline
Why does she think that? & Right? \\\hline
Who do you think the dress belongs to? & Let's try it. \\\hline
And what else? & But you know what? \\\hline
And what were you in your princess's dream? & Just like this. \\\hline
Then I'll puff, and I'll puff, and I'll blow your house down,& So you can  \ldots \\
said the big, bad wolf. & \\\hline
Any clues? & Write your own name on  \ldots \\ \hline
It looks like a little piece of sky in the ground. & Oh, I see. \\\hline
This is something that Dr. King dreamed about. & \ldots that you don't like. So, from now on, \ldots \\\hline
\end{tabular}
\end{center}
\end{table}

To give a qualitative sense of which utteranes the LLM(All) model deemed to be
very high or very low Instructional Support, we list the utterances in the 20-utterance subset in Table \ref{tbl:utterances}.
Ellipses ($\ldots$) denote a sentence
fragment that is likely due to Whisper incorrectly segmenting it from the audio stream.
Among those associated with high CLASS Instructional Support,
we see several questions (``Why does she think that?'', ``So if they're hungry and they want a bottle, how do you think their mommy knows?'', ``What else?'')
that are more open-ended than those in the low Instructional Support list.
Among those associated with \emph{low} Instructional Support, some are very short (``Oh, I see.'', ``Right?''),
and some are more directive (``Just like this.'', ``Oh, maybe you should go look for the letter J.'') than inquisitive.
On the other hand, the utterance ``Aria, give me an animal that was in the story.'' does prompt the student to think
about the story; the fact that this utterance was scored by the model as having a very low $\hat{y}$ could be due to overfitting
of the regression model weights.

\subsection{Hallucations}
We examined qualitatively the set of utterances analyzed in the previous section with a focus on the kinds of LLM hallucations the model would sometimes
make.  First, we modified the \begin{tt}system\end{tt} message to ``\begin{tt}Answer YES or NO and explain the reasoning.\end{tt}'' so that it would
produce more than just a yes/no answer. We then asked Llama2 to judge each of the 100 utterances for each of the 11 binary indicators
listed in Section \ref{sec:indicators}. Based on reading the responses, we noticed three distinct types of hallucations that occurred most frequently:
\begin{itemize}
\item \emph{Incorrect parsing}: Especially for very short input utterances, Llama2 would sometimes misinterpret 
part of the query itself as part of the input sentence it was supposed to analyze. For instance, for the input utterance ``Yes'' (which sometimes
occurred as a single-word utterance),
it turned out that Llama2 incorrectly classified
this utterance as positive for all 11 indicators (i.e., it answered ``YES'' as the first word of its response).
When we asked Llama2 to provide an explanation for one of the indicators (``perform self- and parallel talk''):
\footnotesize
\begin{verbatim}
In the context of a preschool classroom in which a teacher is talking to
their students, does the following sentence "perform self- and parallel
talk" and help students to grow cognitively?
"Yes"
\end{verbatim}
\normalsize
it output:
\footnotesize
\begin{verbatim}
YES.
The sentence "perform self- and parallel talk" is a teaching strategy that
can help preschool students grow cognitively...
\end{verbatim}
\normalsize
It seems that Llama2 interpreted ``perform self- and parallel talk'' as the input utterance of classroom
speech itself, rather than the indicator to inspect it for. This is likely because the actual input
utterance ``Yes'' was short and deemed to be the answer to the query itself.

\item \emph{Subjective misjudgment}: Llama2 sometimes exaggerated the degree to which an utterance
exhibited a CLASS indicator in a way that -- while not completely illogical -- contradicts the CLASS 
standards for what is deemed sufficient for that indicator.
For example, Llama2 judged that the utterance ``Write your own name on'' could ``help students make connections'' because,
``By writing their own names, students can develop a sense of identity and self-awareness''. As another example, Llama2's response to
the input ``Do you want to fix it up?''  was that it ``Develop decision-making skills: By asking the students if they want to fix something, the teacher is giving them the opportunity to make a decision. This helps to develop decision-making skills, which are important for cognitive growth and development.'' Both of these inferences exaggerate the significance of the respective classroom utterances.

\item \emph{Objective misjudgment}: Llama2 would sometimes make objectively false claims about the classroom speech.
For instance, Llama2 judged that the utterance ``Then I'll puff, and I'll puff, and I'll blow your house down, said the big, bad wolf.''
exhibits ``advanced language'' because it contains ``two subordinate clauses ("said the big, bad wolf")''.  In fact, the input contains no such
subordinate clauses. Another example is Llama2's incorrect judgment that ``The sentence "When it was time to do what?" is an example of parallel talk, which is a teaching technique used to encourage cognitive growth in young children.'' This sentence ``When it was time to do what?'' is not parallel talk (describing what a student is doing) in the classroom sense.

\end{itemize}
Finally, we note that we found no instances of total gibberish or of classroom-inappropriate content in Llama2's responses.

There are various strategies that could be used to mitigate these hallucations \cite{tonmoy2024comprehensive}, e.g., prompt engineering, the use of few-shot examples rather
than a zero-shot classification approach, supervised fine-tuning, and many more. We leave these to future work.

%


\section{Visualization \& Explanation of Automated Feedback}
\subsection{Visualization}
\label{sec:visualization}
\begin{figure}
\begin{center}
\includegraphics[width=\columnwidth]{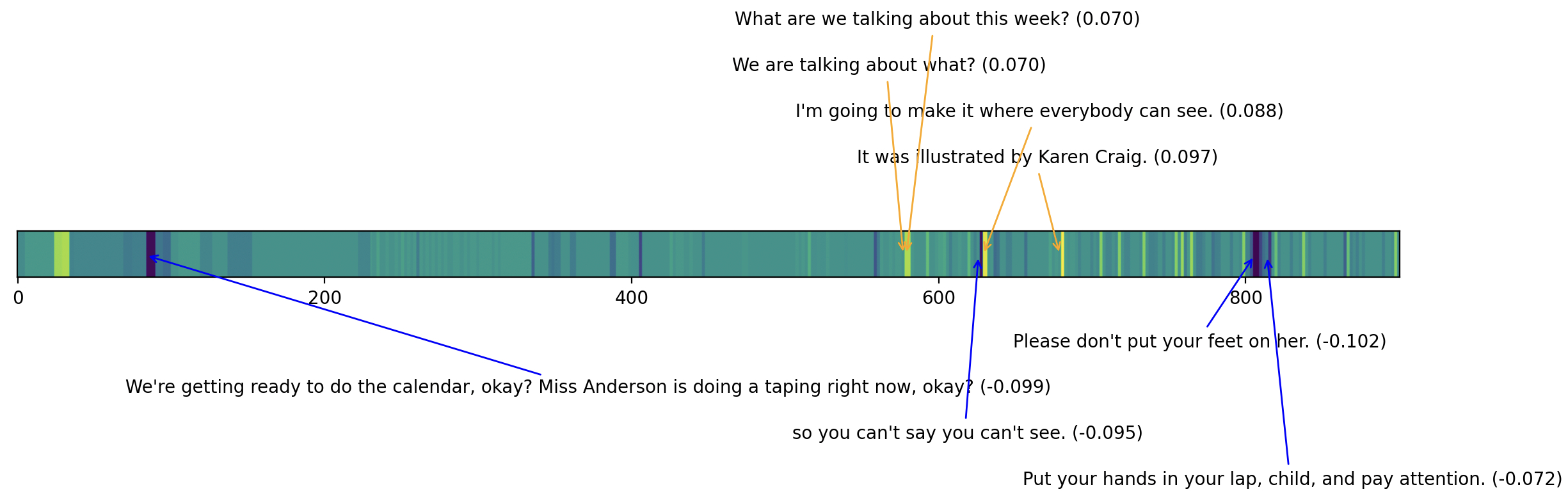}
\end{center}
\caption{Temporal heatmap of CLASS ``Instructional Support'' dimension.
The top 4 most and least highly related utterances from the classroom audio are
highlighted.}
\label{fig:example}
\end{figure}

Given our machine learning approach that is based on analyzing individual utterances
followed by $L_1$-regularized linear regression,
it is straightforward to find the specific utterances in a classroom transcript that contribute the most
toward receiving a high or low CLASS score: We can rank the utterances within an observation session
according to their marginal scores $\Delta \hat{y}^{(i)}$.
An example of this process is given in Figure \ref{fig:example}, where the marginal scores
were computed using the BoW model on the NCRECE PreK dataset: the horizontal axis is time
(in seconds, up to 900sec=15min), and the color value (blue is lowest, orange is highest) at each time $t$ is computed as
a linear function of
$\Delta \hat{y}$ for the utterance that took place at time $t$. The figure additionally shows the 
4 utterances whose $\Delta \hat{y}$ were  highest (above the graph, arrows in orange) and lowest
(below the graph, arrows in blue). Among the negative utterances we see examples of behavioral control
(``Please don't put your feet on her'') rather than intellectual stimulation, and among the positive
utterances we find an open-ended question (``What are we talking about this week?'').

\subsection{Explanation of CLASS Score Estimates}
It is also easy to obtain a simple explanation of a predicted CLASS score. For the BoW model,
an utterance can be characterized as including or not including each of the 302 $n$-grams (including '?' and ' '),
and for the LLM models, by the detected presence/absence of each of the 11 indicators.
Because of the sparsity of the $L_1$ regression coefficients, the number of $n$-grams
with non-zero impact on the marginal CLASS score estimate is typically low.
For example, the utterance ``Please don't put your feet on her.''
received a relatively low $\Delta \hat{y}$ score because it contained the $n$-grams ``please'' and ``put your'',
whose associated standardized regression coefficients were $-0.030$ and $-0.072$.
The utterance ``What are we talking about this week?'' received a relatively high $\Delta \hat{y}$ score because it
contained the $n$-grams ``?'', ``we'', and ``about'', whose associated standardized regression coefficients were
$+0.005$, $+0.053$, and $+0.011$.

The LLM-based models, instead of counting occurrences of $n$-grams, can identify
specific indicators. For instance, the LLM judged that the utterance ``What are we talking about this week?'' exhibits
the indicators ``help students to make connections'', ``provide information'', ``repeat and extend students' language'', and
``repeat and extend students' language'' (note that here we used Llama2 to provide binary answers rather than the real-valued
probability scores).
Moreover, the LLMs can also offer an explanation as to \emph{why} it believes a given utterance exhibits a certain indicator.
By changing the \begin{tt}system\end{tt} message to ``\begin{tt}Answer YES or NO and explain the reasoning.\end{tt}'',
we can ask Llama2 for an explanation of these judgments. For example, with regards to ``help students to make connections'',
it provides useful feedback:
\footnotesize
\begin{verbatim}
The sentence "What are we talking about this week?" can help students to
make connections and grow cognitively in the following ways:
1. Encourages critical thinking: By asking students to think about the topic
of discussion, the teacher is encouraging them to think critically and make
connections between different ideas and concepts.
2. Promotes prior knowledge integration: When students are asked to reflect
on what they are talking about, they are required to integrate their prior
knowledge and experiences with the new information being presented...
...
Overall, the sentence "What are we talking about this week?" can be a
powerful tool for promoting cognitive growth and engagement in a preschool
classroom.
\end{verbatim}
\normalsize

\section{General Discussion}
\subsection{System Accuracy and Implications for Practical Use}
The highest Pearson $R$ attained by any of the automated approaches in Table \ref{tbl:results} was
$0.48$. This level of accuracy is approaching human IRR and is noteworthy
considering that the human raters scored each classroom session from full video,
whereas the automated predictors observed only the text transcript.
Also remarkable is the fact that, although LLMs performed best, the baselines and BoW-based models
were surprisingly competitive on the datasets in our study and are far easier and faster to deploy. On the other hand, these
simplistic approaches arguably capture only superficial features of classroom speech
(similarly to how early methods for automated essay scoring worked \cite{yang2014review}) and could
thus easily be gamed, whereas LLM may have a deeper semantic understanding.


At the current accuracy level ($R=0.48$),
a fully automated system would likely be more useful for educational research rather than for
guidance to individual teachers. In the former context, such a tool might be used to
evaluate the effectiveness of a particular educational intervention in a large-scale study
of hundreds of classrooms (e.g., at the scale of the Measures of Effective Teaching \cite{kane2013have} study),
and the statistical noise in estimating CLASS scores of individual classrooms can be averaged out.
In the latter context (feedback to individual teachers), the accuracy requirements would likely be higher,
and a hybrid human-AI scoring paradigm might be more useful. For instance, the utterance-wise predictions
of an automated predictor could be presented as suggestions rather than
``ground-truth'', and the teachers or coaches could feel free to confirm or refute these suggestions.
This could reduce the labor involved in tedious utterance-by-utterance scoring, provide high accuracy to teachers,
and even help to improve the accuracy of the models by fine-tuning them on the corrected labels.


%
\subsection{Ethics, Equity, and Privacy}
The goal of our research is to harness AI
to give teachers more opportunities to receive specific and useful feedback about their classroom discourse
as well as to make classroom observation scoring more efficient and accurate for large-scale research use.
AI-based systems could be deployed at the teacher's discretion, without the need to share the
results with anyone else. Since the methods described in this paper used only classroom audio and no video,
the privacy infringement is reduced significantly. Moreover, since locally-executable LLMs such as Llama2 can 
be run on a school computer, there is no need to upload a classroom video to a private company for processing.

AI-based classroom observation has the potential to promote educational equity by giving 
detailed feedback to teachers who otherwise could not receive it due to geographical distance, financial constraints, additional family responsibilities,
etc. \cite{lesiak2021turning}.  To realize this potential, it is necessary to train such models
on diverse classroom datasets. Moreover, before deploying them at scale, it is important to evaluate them for
potential biases using appropriate accuracy metrics (e.g., ABROCA \cite{gardner2019evaluating}).

\section{Conclusions}
We have explored how either Large Language Models  or classic Bag of Words  models, can, with the right scaffolding and task decomposition,
be used to estimate the level of ``Instructional Support'' of a classroom, given an automatically generated transcript
of the classroom speech. In particular, we proposed a machine learning architecture whereby a transcript
is analyzed at the utterance level for specific behavioral indicators associated with the Instructional Support
domain  of the CLassroom Assessment Scoring System (CLASS), and then these utterance-wise judgments are aggregated
using $L_1$-regularized linear regression to produce a global score estimate. We conducted
experiments on two CLASS-coded datasets of toddler and pre-kindergarten classrooms to validate the approach.
The results suggest that, at the global CLASS score level,  the
system's accuracy is similar to human inter-rater reliability. Moreover, since the global estimates are grounded in utterance-level
judgments, it can give teachers specific and explainable feedback about which utterances were positively/negatively correlated  with Instructional Support. 
Importantly, this automated approach comes at a substantial savings to the education research and support field. While there is still substantial room for improvement by the automated approaches, particularly
at the individual utterance level,  AI-assisted coding shows substantial promise as a step forward in providing timely and accurate feedback to the education field on practices known to foster student development. Thus, building on this approach -- in terms of accuracy, equity, and the ways in which the feedback can be user-friendly to individuals -- has the potential to positively shape the field of classroom observation for the better.  

\subsection{Limitations}
\label{sec:limitations}
\emph{Automatic speech recognition}: The entrypoint to our CLASS score estimation system
is the automatic speech recognition provided by Whisper. It is possible that the speech recognition 
accuracy differs across demographics such as gender and ethnicity \cite{koenecke2020racial}. This could result in
accuracy disparities of downstream CLASS score estimates as well.

\emph{Interpretation of cross-validation accuracy}: We followed a standard cross-validation approach
whereby we partitioned each dataset (UVA Toddler, NCRECE PreK) into 5 disjoint folds, stratified such that
(a) no teacher appeared in more than one fold and (b) each of the 5 folds received a similar
distribution of CLASS scores. Because each of the 5 testing folds contains multiple teachers,
and because there is significant variance in CLASS scores across teachers,
the cross-validation accuracy captures how well the trained model can \emph{distinguish among
teachers} those who tend to receive higher CLASS scores from those who tend to receive lower
CLASS scores.  Note that this contrasts with leave-one-teacher-out cross-validation, which would express
how well the trained model can distinguish classroom sessions with low from high CLASS
scores \emph{within multiple classrooms from the same teacher}.

\emph{CLASS behavioral indicators and non-negative regression coefficients}:
Because our goal was to explore the extent
to which the proposed techniques can estimate CLASS scores, we applied standard $L_1$-regularized linear regression
for prediction, which allows the learned coefficients to be any real number. However, according
to the CLASS Manual \cite{pianta2008classroom}, each CLASS dimension is scored based on the
\emph{positive evidence} that it exists;  it should not be judged based on observed behaviors or events
that are negatively correlated with the target dimension. For example, just because a teacher
refrains from asking \emph{closed}-ended questions does not mean that they should receive a high score for Language Modeling.
To enforce the constraint that all regression coefficients 
should be non-negative,
a  technique such as non-negative $L_1$-regularized linear regression \cite{ang2020}
can be used. See Appendix for details.

\emph{Statistical assumptions of linear regression}:
The primary focus of our paper is on the predictive accuracy of the proposed models for CLASS score estimation, not
the estimation of the regression coefficients themselves. For visualizing and explaining the model's predictions
to teachers (Section \ref{sec:visualization}), however, the reliability of these coefficient estimates does come into play:
These coefficients
represent how the presence of one of the behavioral indicators (as estimated by the LLM), or how the count of
a particular $n$-gram, impacts the final CLASS score prediction. Unbiased estimation of the linear
regression coefficients requires that the standard assumptions of linear regression
(linearity, homoscedasticity, etc.) hold true. Given that our models are high-dimensional
(e.g., we used 302 BoW features), and given that checks for linearity are usually conducted manually (and often subjectively)
by inspecting scatter plots, this would be infeasible in our case. Hence, the quantitative estimates of how
individual utterances contribute to the final CLASS score prediction for a given observation session should be treated with
caution.

\emph{Limitations of existing observation protocols}: 
Even widely-used protocols such as the CLASS have only modest short-term associations \cite{burchinal2018measuring}
with students' learning. More work is needed to elucidate the factors that impact children \cite{burchinal2020does,pianta2020children}.
New AI-based approaches may be useful for this purpose.


\subsection{Future Work}
Future work on AI for classroom observation 
can investigate how the LLM-based judgments of individual utterances can benefit from global information from the entire classroom transcript.
Even more powerfully,
multi-modal LLMs (e.g., NExT-GPT \cite{wu2023nextgpt}) could analyze not just the transcript
but the audio and video as well, e.g., to infer more about the educational and emotional context,
or to analyze which kinds of classroom activities (meals, read-aloud, etc.)
are associated with different levels of interaction.
Moreover, it would be valuable to explore whether AI-based 
tools can predict  other classroom observation measures beyond Instructional Support, and, more generally,
whether they can predict students' downstream learning outcomes, not just classroom observation
scores from human coders.
Finally, by tracking the dialogue from teachers to \emph{individual} students,
it would be possible to analyze  classroom speech from an equity perspective.

\subsection*{Acknowledgement}
This research was supported by the NSF National AI Institute for Student-AI Teaming (iSAT) under grant DRL \#2019805, and also from an NSF CAREER grant \#2046505.  The opinions expressed are those of the authors and do not represent views of the NSF.

\bibliographystyle{acmtrans}
\bibliography{paper}

\section*{Appendix}
\subsection*{NCRECE Subset}
Due to the significant manual effort that was necessary to convert the old digital video (DV) tapes on which
NCRECE was originally collected into a format that was usable for our analyses, we used only a subset
of the entire NCRECE dataset.
Compared to the entire dataset, the subset's distribution of CLASS scores
was statistically significantly different for the Language Modeling dimension ($\chi^2(6)=12.892, p=0.045$). However, the
actual difference in the probability distributions over Language Modeling scores (1-7) was small:
\\
$[0.16132479, 0.40598291, 0.24679487, 0.1207265,  0.04380342, 0.01923077, 0.00213675]$ (subset)
versus\\
$[0.16824275, 0.35383976, 0.27008209, 0.13041671, 0.05902525, 0.01662683, 0.0017666]$ (whole).

There was no statistically significant difference found between the subset and whole-sample scores for
either the Quality of Feedback ($\chi^2(6)=4.128, p=0.659$) or the Concept Development dimensions ($\chi^2(6)=0.660, p=0.995$).

\subsection*{Non-Negative $L_1$-Regularized Linear Regression}
We explored an approach
based on non-negative $L_1$-regularized linear regression \cite{ang2020} which, in addition to sparsity,
requires all coefficients to be non-negative. We found
slightly worse performance of this approach compared to unconstrained $L_1$ regression.
On UVA Toddler, the best model was the LLM(QualFbk), which 
achieved a Pearson correlation of $0.37$, and on
NCRECE, it was the LLM(All) model, which achieved a Pearson correlation of $0.43$.
%

\subsection*{$n$-grams}
We applied a stop-word list consisting of:

\footnotesize
[
'a', 'an', 'and', 'are', 'as', 'at', 'be', 'but', 'by', 'for', 'if', 'in', 'into', 'is', 'it', 'no', 'not', 'of', 'on', 'or', 'such', 'that', 'the', 'their', 'then', 'there', 'these', 'they', 'this', 'to', 'was', 'will', 'with'].
\normalsize

After removing stop-words, the 300 most frequently occurring $n$-grams in the UVA Toddler dataset were
(in decreasing order of frequency):

\footnotesize
\noindent ['you', 'i', 'your', 'go', 'do', 'can', 'what', 'good', 'come', 'going', 'all', 'here', 'me', 'okay', 'going to', 'we', 'right', 'want', 'have', 'up', 'put', "it's", 'sit', 'thank', 'thank you', 'see', 'one', "let's", 'job', 'oh', 'do you', "don't", 'down', "i'm", 'good job', 'so', 'get', 'you want', 'yeah', 'like', 'want to', 'come on', 'all right', 'are you', "that's", "we're", 'my', "you're", 'look', 'can you', 'know', 'on the', 'yes', 'back', 'you want to', 'got', 'in the', 'how', 'did', 'sit down', 'okay?', 'he', 'over', 'our', 'please', 'say', 'you can', 'let', "he's", 'put it', 'hands', 'now', "we're going", "what's", 'ready?', "we're going to", 'do you want', "i don't", 'some', "i'm going", 'it?', "i'm going to", 'little', 'you have', 'out', 'them', 'have a', 'two', 'you go', 'color', 'need', 'more', 'just', 'guys', 'turn', 'let me', 'morning', 'she', 'to the', 'that?', 'to go', 'his', 'play', 'take', 'come here', 'give', 'eat', 'red', 'where', 'have to', 'good morning', 'baby', 'why', 'ready', 'time', 'her', 'what is', 'do you want to', "don't know", 'on your', 'make', 'to do', "i don't know", 'in your', 'wait', 'does', 'did you', 'look at', 'three', 'think', 'hold', 'very', 'you are', "let's go", 'jump', 'ball', 'your hands', 'too', 'friends', "where's", 'you guys', 'green', 'find', 'i want', "i'll", 'hey', 'big', 'this?', 'at me', 'it in', 'what color', 'alright', 'five', 'when', 'blue', 'about', 'what do', 'table', 'hi', 'nice', 'me see', 'let me see', 'book', 'over here', 'way', 'seat', 'this is', 'clean', 'put your', 'high', 'to get', 'done', 'him', 'wash', 'there you', 'mad', 'off', 'right here', 'there you go', 'yellow', 'we have', 'around', "why he's", 'you got', 'who', 'i like', 'mad at', "can't", 'what you', 'mad at me', 'what do you', 'me i', '289', 'everybody', "there's", 'love', 'to be', 'to put', 'four', 'know why', 'sing', 'no no', "me i don't", '288', 'today', 'open', 'i know', 'mommy', 'help', 'at the', 'go to', 'at me i', 'do it', 'you need', 'said', "he's mad", "why he's mad", 'you like', 'see you', "at me i don't", 'miss', 'has', 'you know', 'i have', 'mad at me i', 'if you', 'chair', "know why he's", "he's mad at", "why he's mad at", "it's a", 'world', "mad at me i don't", "don't know why", "he's mad at me", "why he's mad at me", "i don't know why", 'put it in', 'you ready?', "me i don't know", 'wonder', 'because', 'feet', 'bus', "at me i don't know", 'need to', 'use', 'what?', 'how i', 'read', "let's see", 'you see', 'the table', 'i wonder', 'the world', "don't know why he's", 'outside', 'you?', 'above', "i don't know why he's", 'yay!', 'you put', 'i see', 'how i wonder', 'what you are', "he's mad at me i", 'is it', 'is this?', 'walk', 'circle', 'day', "know why he's mad", 'stand', 'try', 'up above', 'color is', 'got to', "know why he's mad at", 'can i', 'what color is', 'i think', 'all the', 'hand', 'away', 'one two', "you don't", 'right?', "me i don't know why", 'a little', 'and then', "don't know why he's mad", 'put the', 'to sit', 'uh-oh', 'mouth', 'purple', 'milk', 'song', 'one?', 'show'].
\normalsize

In NCRECE PreK, they were:

\footnotesize
\noindent ['you', 'i', 'what', 'your', 'do', 'we', 'going', 'going to', 'okay', 'can', 'have', 'so', 'good', 'see', 'he', 'all', 'me', 'right', 'go', 'do you', 'up', 'one', "it's", 'like', "let's", 'my', "i'm", 'here', 'put', 'did', "that's", 'our', 'in the', "we're", 'think', 'how', 'want', 'now', 'she', 'know', "you're", 'look', 'about', 'get', 'letter', 'very', 'down', "don't", 'little', 'yes', 'them', 'his', 'when', 'said', 'on the', 'oh', 'because', 'say', 'out', 'you think', 'some', "we're going", "we're going to", 'you can', 'job', 'need', 'come', 'just', 'name', 'this is', 'can you', 'two', 'want to', 'make', "i'm going", 'what do', "i'm going to", 'thank', 'her', 'thank you', 'does', 'have a', 'who', 'very good', 'good job', 'all right', 'back', 'sit', 'yeah', 'got', "what's", 'what?', 'friends', 'where', 'to the', 'do you think', 'over', 'what is', 'hands', 'to do', 'what do you', 'okay?', 'of the', 'today', 'would', 'you have', 'let', 'has', 'you know', 'tell', 'are you', 'look at', 'three', 'word', 'we have', 'if you', "let's see", 'i want', 'read', "he's", 'time', 'too', 'everybody', 'give', 'to be', 'have to', 'big', 'story', 'book', 'well', 'you want', 'turn', 'hand', 'and then', 'it?', 'a little', 'morning', 'take', 'i have', 'sound', 'why', 'and the', 'is the', 'more', 'us', 'to go', 'first', 'ready', 'this?', 'remember', 'did you', 'him', 'is a', 'please', 'picture', 'way', 'the letter', 's', 'had', "they're", 'from', 'something', 'at the', 'find', 'let me', "it's a", 'were', 'help', 'next', 'red', 'on your', 'around', 'show', 'hear', 'water', 'four', 'write', 'you to', 'color', 'with the', "i don't", "can't", 'five', 'guys', 'that?', "there's", 'words', 'listen', 'go to', 'your hands', 'you want to', 'hold', 'cat', 'another', 'need to', 'and you', 'baby', 'do?', 'stand', 'else', "you're going", 'to get', 'to put', 'day', 'says', 'again', 'in your', "you're going to", 'start', 'you see', 'eat', "didn't", 'those', 'ready?', 'things', 'use', 'bear', 'what letter', 'all the', 'friend', 'right here', "i'll", 'i need', 'good morning', 'play', 'what did', 'm', 'really', 'put it', 'many', 'n', 'your name', 'put your', 'to make', 'doing', 'going to do', 'green', 'went', 'move', 'blue', 'is this?', 'in a', 'to see', 'wait', 'i can', 'and what', "she's", 'kind', 'house', 'you guys', 'called', 'r', 'tell me', 'do we', 'like to', 'and i', 'what do you think', 'other', 'it is', 'could', 'l', 'it was', 'i see', 'off', 'kind of', 'and a', 'your hand', 'sit down', 'clap', 'people', 'try', 'stand up', 'e', 'dog', 'know what', 'i like', 'we can', 'd', 'table', 'going to be', 'i know', 'alright', 'up and', 'b', 'shake', 'you go', 'come on', 'wow', 'talk', 'how do', 'together', 'after', 'letter?', 'snow', 'is this', 'thing', "you don't", 'how many', 'what are', 'with a'].
\normalsize

\subsection*{LLM Prompts}
For the UVA Toddler dataset (Toddler CLASS), we asked Llama2 to examine the following indicators:
(1) ``provide active facilitation of children's learning'', 
(2) ``expand children's cognition'',
(3) ``promote children's active engagement'',
(4) ``provide scaffolding",
(5) ``provide information",
(6) ``encourage and affirms",
(7) ``ask open-ended questions",
(8) ``repeat and extend students' language",
(9) ``perform self- and parallel talk", and
(10) ``use advanced language".

\subsection*{Alternative LLM Approaches}
In addition to the zero-shot prompting approach, we explored several approaches that did not work as well:
(1) We tried automated prompt engineering whereby we asked Llama2 to generated hundreds of semantically equivalent
variations of the prompts shown above; then, we used a Gaussian Process (GP) \cite{williams2006gaussian}
to optimize (on a small subset of our datasets) the input prompt in terms of its CLASS score estimation accuracy. For
this purpose, we defined the covariance function of the GP to be the squared-exponential function of the $L_2$ distance between
the embedding vectors of the two prompts, as given by Sentence-BERT \cite{reimers2019sentence}.
However, our pilot experiments suggested that all the generated prompts perform about equally well for CLASS score prediction,
and we thus abandoned the idea. (2) We also
tried fine-tuning Llama2 on long-form text (the entire classroom transcript of each 15-minute session) using
Low-Rank Adaptation \cite{hu2021lora} but found that the resulting model usually just output the median CLASS score of the training labels.


\end{document}